# Persian Slang Text Conversion to Formal and Deep Learning of Persian Short Texts on Social Media for Sentiment Classification

***M. Khazeni, M. Heydari*, A. Albadvi***
*Department of IT Engineering, Faculty of Industrial and Systems Engineering, Tarbiat Modares University, Tehran, Iran.*

*Corresponding Author's Email Address:
*m_heydari@modares.ac.ir*

## Abstract

**Background and Objectives:** The lack of a suitable tool for the analysis of conversational texts in Persian language has made various analyzes of these texts, including Sentiment Analysis, difficult. In this research, it has we tried to make the understanding of these texts easier for the machine by providing PSC, Persian Slang Convertor, a tool for converting conversational texts into formal ones, and by using the most up-to-date and best deep learning methods along with the PSC, the sentiment learning of short Persian language texts for the machine in a better way.

**Methods:** Be made More than 10 million unlabeled texts from various social networks and movie subtitles (as dialogue texts) and about 10 million news texts (as official texts) have been used for training unsupervised models and formal implementation of the tool. 60,000 texts from the comments of Instagram social network users with positive, negative, and neutral labels are considered as supervised data for training the emotion classification model of short texts. The latest methods such as LSTM, CNN, BERT, ELMo, and deep processing techniques such as learning rate decay, regularization, and dropout have been used. LSTM has been utilized in research, and the best accuracy has been achieved using this method.

**Results:** Using the official tool, 57% of the words of the corpus of conversation were converted. Finally, by using the formalizer, FastText model and deep LSTM network, the accuracy of 81.91 was obtained on the test data.

**Conclusion:** In this research, an attempt was made to pre-train models using unlabeled data, and in some cases, existing pre-trained models such as ParsBERT were used. Then, a model was implemented to classify the Sentiment of Persian short texts using labeled data.

## Introduction

With the increasing accessibility of digital platforms, such as websites and social networks, there is a continual surge in the volume of data generated from these sources. This data serves as a valuable resource for managers and researchers, facilitating diverse analyses. An aspect worthy of examination involves the emotional content within texts, a task traditionally performed manually by human resources in the past. Contemporary advancements enable the delegation of this analysis to machines, handling larger datasets with heightened efficiency, precision, and comprehensiveness. Nevertheless, data sourced from social networks is inherently unstructured and characterized by its complexities. This research endeavors to introduce a system designed for the analysis of conversational and concise texts originating from Persian language social networks. The ensuing section refers to the delineation, objectives, significance, and historical context of the subject matter. Emphasizing the analysis of text emotions offers multifaceted applications across commercial, economic, educational, political, and cultural domains. In recent years, a spectrum of methodologies leveraging text mining algorithms, natural language processing, and emotion dictionaries has emerged for emotion analysis and opinion mining. Notably, the proliferation of deep neural networks has gained prominence in this domain, reflecting their computational prowess, akin to their widespread adoption in various analytical fields.

## Related Works

Within the sentiment analysis domain of short texts in Persian, several research initiatives have been executed. Nevertheless, none of these investigations have relied on extensive and dependable conversational datasets, thus lacking requisite comprehensiveness and confidence. Additionally, the historical trajectory of studies concerning Persian conversational language extends considerably, predominantly centering on the elucidation of features, characteristics, and rules inherent to spoken Persian. However, a limited number have explored the conversion of slang texts into formal language. This study stands as the inaugural substantive exploration of transforming colloquial texts into formalized expressions, addressing a notable gap in existing research endeavors [1].

The progression of natural language processing and text mining has spurred a surge in research devoted to refining sentiment analysis methods. The primary objective is to adeptly handle expansive datasets with enhanced precision and efficiency. Recognizing the importance of sentiment analysis and acknowledging the distinct challenges posed by both formal and conversational nuances within the Persian language, there arises a pressing demand for innovative models and methodologies to effectively scrutinize sentiments within extensive textual datasets originating in Persian [2].

The diverse applications and lucrative implications of sentiment analysis, coupled with the linguistic intricacies presented by the Persian language, underscore a significant and pressing issue. Addressing this matter necessitates foundational solutions derived from comprehensive and extensive studies within this domain [3].

A prevalent concern within text mining pertains to sentiment analysis, also known as opinion mining. This process entails the computational scrutiny of opinions, evaluations, attitudes, and emotions conveyed by individuals regarding various entities, such as individuals, issues, events, topics, and their respective attributes [4].

In nearly all sources, sentiment analysis is commonly examined at three primary levels: document, sentence, and aspect. At the document level, the goal is to determine whether the overall sentiment of the document is positive or negative. At the sentence level, the sentiment is assessed for each individual sentence. Finally, at the aspect level, the sentiment is investigated for each feature or entity mentioned within the sentence [5].

Challenges encountered in machine learning for short texts include a) Brevity: Short texts, comprising only a few words, may lead to insufficient representation of the document for comprehension or learning. b) Feature Limitations: Short texts have limited length, and this restricted capacity must be utilized to express diverse topics for users, each using their own vocabulary and writing style. Therefore, a specific topic may have diverse content, making it challenging to extract precise features from short texts. c)Swift Processing: Due to its practical application, a short text needs to be processed very quickly, and results need to be conveyed promptly. d) Spelling Errors and Informal Writing: In many cases, especially in opinions expressed on microblogs and social networks, a short text is summarized, resulting in numerous spelling errors, informal writing, or colloquial expressions [6].

Naemi et al., addressed conversion of Persian informal words to formal words by using the spell-checking approach. They extracted two datasets included formal and informal words from the four most visited news websites in Persian. Results show that their proposed system can detect approximately 94% of the Persian informal words, with the ability to detect 85% of the best equivalent formal words [7].

Tajalli et al., building a parallel corpus of 50,000 sentence pairs with alignments in the word/phrase level. The sentences were attempted to cover almost all kinds of lexical and syntactic changes between informal and formal Persian, therefore both methods of exploring and collecting from the different resources of informal scripts and following the phonological and morphological patterns of changes were applied to find as much instances as possible. The corpus has about 530,000 alignments and a dictionary containing 49,397 word and phrase pairs [8].

Rasooli et al., proposed an effective standardization approach based on sequence-to-sequence translation. They designed an algorithm for generating artificial parallel colloquial-to-standard data for learning a sequence-to-sequence model and annotated publicly available evaluation data consisting of 1912 sentences. Their model improves English-to-Persian machine translation in scenarios for which the training data is from colloquial Persian with 1.4 absolute BLEU score difference in the development data, and 0.8 in the test data [9].

Mazoochi et al., constructed a user opinion dataset called ITRC-Opinion in a collaborative environment and insource way contained 60,000 informal and colloquial

Persian texts from social microblogs such as Twitter and Instagram. They proposed a new architecture based on the convolutional neural network (CNN) model for more effective sentiment analysis of colloquial text in social microblog posts. The constructed datasets are used to evaluate the presented architecture. Some models, such as LSTM, CNN-RNN, BiLSTM, and BiGRU with different word embeddings, including FastText, Glove, and Word2vec, investigated our dataset and evaluated their results. Their model reached 72% accuracy [10].

Momtazi et al., explored the issue of informal texts, and suggested a framework for transforming informal texts into formal texts in the Persian language. Two cutting-edge sequence-to-sequence models, specifically the encoder-decoder and transformer-based models, are employed for this purpose. Alongside neural network models, a series of guidelines for converting informal text to formal text are introduced, and the effects of integrating these guidelines with each of the two models are analyzed. The evaluation of their proposed frameworks reveals that the optimal performance, with an accuracy of 70.7% in the SacreBLEU metric, is achieved through the utilization of the transformer-based model in conjunction with the set of guidelines [11].

Nezhad et al., investigations on sarcasm detection technique in Persian tweets was examined through the integration of various machine learning and deep learning approaches. A series of feature sets encompassing diverse forms of sarcasm were introduced, specifically deep polarity, sentiment, part of speech, and punctuation features. These features were employed for the categorization of tweets into sarcastic and nonsarcastic categories. The deep polarity feature was formulated by executing a sentiment analysis utilizing a deep neural network architecture. Moreover, a Persian sentiment lexicon comprising four sentiment classifications was constructed to extract the sentiment feature. Additionally, a novel Persian proverb lexicon was incorporated during the preparatory phase to enhance the precision of the proposed model. The model's performance was assessed through a range of standard machine learning techniques. The experimental outcomes demonstrated that the method surpassed the baseline approach, achieving an accuracy rate of 80.82%. The research also delved into the significance of each feature set proposed and appraised their contribution to the classification process [12].

Golazizian et al., in their research, which is the first attempt at irony detection in Persian language, emoji prediction is used to build a pretrained model. The model is finetuned utilizing a set of hand labeled tweets with irony tags. A bidirectional LSTM (BiLSTM) network is employed as the basis of our model which is improved by attention mechanism. Additionally, a Persian corpus for irony detection containing 4339 manually labeled tweets is introduced. Experiments show the proposed approach outperforms the adapted state-of-the-art method tested on Persian dataset with an accuracy of 83.1% and offers a strong baseline for further research in Persian language [13].

Hajiabdollah et al., explored on Improving polarity identification in sentiment analysis using sarcasm detection and machine learning algorithms in Persian tweets. To accomplish their study, 8000 Persian tweets that have emotional labels and examined for the presence or absence of sarcasm have been used. The innovation of their research is in extracting keywords from sarcastic sentences. In their research, a separate classifier has been trained to identify irony of the text. The output of this classifier is provided as an added feature to the text recognition classifier. In addition to other keywords extracted from the text, emoticons and hashtags have also been used as features. Naive Bayes, support vector machines, and neural networks were used as baseline classifiers, and finally the combination of classifiers was used to identify the feeling of the text. The results of this study show that identifying the irony in the text and using it to identify emotions increases the accuracy of the results [14].

Najafi-Lapavandani et al., investigated on Humor Detection in Persian. As one of the early efforts for detecting humor in Persian, their research proposes a model by fine-tuning a transformer-based language model on a Persian humor detection dataset. The proposed model has an accuracy of 84.7% on the test set. Moreover, their research introduced a dataset of 14,946 automatically-labeled tweets for humor detection in Persian [15].

Sharma et al. proposed a model for classifying short sentimental sentences using a CNN-enhanced with fine-tuned Word2Vec embeddings. Their approach demonstrated improved classification performance, highlighting the efficacy of CNNs in handling short text sentiment analysis [16].

Muhammad et al. conducted sentiment analysis on Indonesian hotel reviews utilizing a combination of Word2Vec embeddings and LSTM networks. Their study achieved significant improvements in sentiment classification accuracy, underscoring the effectiveness of integrating Word2Vec with LSTM for capturing contextual information in text [17].

Ouchene et al. explored sentiment analysis on Algerian tweets using FastText embeddings combined with LSTM networks. Their empirical study demonstrated that this hybrid approach effectively captured sentiment nuances in Algerian Arabic tweets, offering a robust methodology for sentiment analysis in low-resource languages [18].

Patel et al. introduced a hybrid deep learning approach for rumor detection, combining ELMo embeddings with CNN. This model achieved enhanced accuracy in detecting rumors, leveraging the contextual embeddings from ELMo and the feature extraction capabilities of CNNs to address the complexities of rumor detection in text [19].

Farahani et al. developed ParsBERT, a transformer-based model tailored for Persian language understanding. ParsBERT significantly advanced the state of natural language processing in Persian, providing a powerful tool for various downstream tasks such as sentiment analysis, text classification, and named entity recognition [20].

Pires et al. investigated the multilingual capabilities of BERT, analyzing its performance across multiple languages. Their findings revealed that multilingual BERT can effectively handle a variety of languages, including those with limited resources, making it a versatile tool for multilingual natural language processing tasks [21].

Alkhalifa et al., addresses the challenge of humor detection in natural language processing, particularly for Arabic, a language with limited resources. The authors collected and annotated humorous tweets in both Arabic dialects and Modern Standard Arabic (MSA). They evaluated seven Arabic pre-trained language models (PLMs)—AraBERTv02, Arabertv02-twitter, QARIB, MarBERT, MARBERTv2, CAMeLBERT-DA, and CAMeLBERT-MIX—by fine-tuning them on this dataset. The results indicated that CAMeLBERT-DA performed best, achieving an F1-score and accuracy of 72.11% [22].

Eke et al., tackles the challenge of sarcasm detection in natural language processing by proposing a context-based feature technique using both deep learning and conventional machine learning models. Traditional models often focus solely on content, neglecting contextual information and sentiment polarity, leading to ineffective sarcasm detection. The study introduces three models: (1) a deep learning model with Bi-LSTM and GloVe embeddings for context learning, (2) a Transformer-based model using the BERT architecture, and (3) a feature fusion model combining BERT, sentiment-related features, syntactic features, and GloVe embeddings with conventional machine learning. Evaluations on Twitter and Internet Argument Corpus (IAC-v2) datasets show high precision rates of 98.5% and 98.0%, demonstrating the effectiveness of the proposed approach [23].

Shatnawi et al., presented BFHumor, a BERT-Flair-based humor detection model designed to identify humor in news headlines. The model combines several state-of-the-art pre-trained NLP techniques in an ensemble approach. Evaluated using SemEval-2020 public humor datasets, BFHumor achieved notable results with a Root Mean Squared Error (RMSE) of 0.51966 and an accuracy of 0.62291. The study also explores the reasons for the model's effectiveness through experiments on the BERT model, revealing that BERT captures surface knowledge in lower layers, syntactic features in middle layers, and semantic understanding in higher layers [24].

Annamoradnejad et al., introduces a novel approach for detecting and rating humor in short texts, leveraging a well-known linguistic theory of humor. The method involves separating sentences within a text, generating embeddings using the BERT model, and feeding these embeddings into a neural network to analyze congruity and latent relationships between sentences for humor prediction. The approach is validated using a newly created dataset of 200,000 labeled short texts for binary humor detection. Additionally, the model was tested in a live machine-learning competition on Spanish tweets, achieving F1 scores of 0.982 and 0.869. These results outperform both general and state-of-the-art models. The study highlights that the effectiveness of the model is significantly attributed to the use of sentence embeddings and the incorporation of humor's linguistic structure in the model design [25].

Sadjadi et al., addresses the challenge of measuring semantic similarity in Persian informal texts, which has been poorly served by previous methods. Traditional approaches have struggled with both accuracy and handling colloquial language. To overcome these limitations, the study introduces a new transformer-based model, FarSSiBERT, specifically designed for Persian informal short texts from social networks. This model is built using the BERT architecture, trained from scratch on approximately 104 million Persian informal texts, and supported by a novel tokenizer that effectively handles informal language. Additionally, a new dataset, FarSSiM, has been created with real social network data and annotated by linguistic experts. The FarSSiBERT model outperforms existing models like ParsBERT, laBSE, and multilingual BERT in measuring semantic similarity, and shows promise for broader NLP tasks involving colloquial Persian text and informal tokenization [26].

Falakaflaki et al., addresses the challenge of formality style transfer in Persian, a task complicated by the growing use of informal language on digital platforms. The goal is to convert informal text into formal text while preserving its original meaning, considering both lexical and syntactic differences. The authors propose a new model, Fa-BERT2BERT, which builds on the Fa-BERT architecture and integrates consistency learning with gradient-based dynamic weighting. This model enhances understanding of syntactic variations and improves balance in loss components during training. Evaluation against existing methods using new metrics tailored to syntactic and stylistic changes shows Fa-BERT2BERT's superior performance across BLEU, BERT score, Rouge-l, and other metrics. This advancement enhances Persian language processing by improving the accuracy and functionality of NLP tools, which can streamline content moderation, enhance data mining, and support effective cross-cultural communication [27].

Dashti et al., presented an advanced Persian spelling correction system that integrates deep learning with phonetic analysis to improve accuracy and efficiency in NLP. The system employs a fine-tuned language representation model to combine deep contextual understanding with phonetic insights, effectively addressing both non-word and real-word spelling errors. It is particularly adept at handling the complexities of Persian spelling, such as its intricate morphology and homophony. Evaluations on a comprehensive dataset reveal the system's exceptional performance, with F1-Scores of 0.890 for detecting real-word errors, 0.905 for correcting them, and 0.891 for non-word error correction. These results demonstrate the effectiveness of incorporating phonetic analysis into deep learning models for spelling correction, advancing Persian language processing and highlighting a valuable approach for future research in the field [28].

Kebriaei et al., addresses the issue of hate and offensive language on social networks, focusing on Twitter and Persian language content. Due to the scarcity of resources for Persian, the researchers compiled a dataset of 38,000 Persian tweets containing hate and offensive language, using keyword-based selection and crowdsourced lexicons. The dataset includes a Persian offensive lexicon and nine target-group lexicons. Manual annotation was performed by multiple annotators to ensure accuracy. The study also evaluated potential biases in the dataset using two assessment criteria (FPED and pAUCED) and adjusted the dataset to reduce bias. The results show that while bias was significantly reduced, the F1 score was minimally affected, demonstrating the effectiveness of the bias mitigation strategy [29].

Vakili et al., explores advanced sentiment analysis techniques for Persian Twitter content by combining multiple approaches: the Naive Bayes classifier, a custom rule-based model, and the BERT transformer model. While traditional models like SVM, Naive Bayes, and MLP show limitations in isolation, the hybrid model developed in this study integrates these methods and achieves notable improvements. The hybrid approach, which combines Naive Bayes and a bespoke rule-based model with BERT, outperforms BERT alone, reaching an accuracy of 89% compared to BERT's 86%. Despite being slightly more complex, this hybrid model maintains comparable computational efficiency to BERT fine-tuning and enhances sentiment classification effectiveness for social media applications [30].

Table 1: Related works comparison

| Study | Objective | Dataset | Method | Results |
|---|---|---|---|---|
| **Naemi** | Conversion of Persian informal words to formal words | Formal and informal word datasets from Persian news websites | Spell-checking approach | 94% detection of informal words; 85% detection of formal equivalents |
| **Tajalli** | Building a parallel corpus for informal-to-formal Persian text transformation | 50,000 sentence pairs with 530,000 alignments and a dictionary of 49,397 pairs | Resource collection and phonological/morphological pattern analysis | Comprehensive corpus and dictionary for informal-to-formal conversion |
| **Rasooli** | Standardization of colloquial to formal Persian using sequence-to-sequence translation | 1,912 annotated sentences | Sequence-to-sequence model for artificial parallel data generation and standardization | 1.4 BLEU score improvement for English-to-Persian translation |
| **Mazoochi** | Sentiment analysis of colloquial Persian texts | 60,000 informal and colloquial Persian texts from microblogs | CNN-based architecture with various embeddings | Achieved 72% accuracy in sentiment analysis |
| **Naemi** | Conversion of Persian informal words to formal words | Formal and informal word datasets from Persian news websites | Spell-checking approach | 94% detection of informal words; 85% detection of formal equivalents |
| **Tajalli** | Building a parallel corpus for informal-to-formal Persian text transformation | 50,000 sentence pairs with 530,000 alignments and a dictionary of 49,397 pairs | Resource collection and phonological/morphological pattern analysis | Comprehensive corpus and dictionary for informal-to-formal conversion |
| **Rasooli** | Standardization of colloquial to formal Persian using sequence-to-sequence translation | 1,912 annotated sentences | Sequence-to-sequence model for artificial parallel data generation and standardization | 1.4 BLEU score improvement for English-to-Persian translation |
| **Mazoochi** | Sentiment analysis of colloquial Persian texts | 60,000 informal and colloquial Persian texts from microblogs | CNN-based architecture with various embeddings | Achieved 72% accuracy in sentiment analysis |

| Study | Objective | Dataset | Method | Results |
|---|---|---|---|---|
| **Momtazi** | Formalization of informal Persian texts using sequence-to-sequence and transformer models | No specific dataset mentioned, evaluation on existing data | Encoder-decoder and transformer models with integration guidelines | 70.7% accuracy using transformer-based model with guidelines |
| **Nezhad** | Sarcasm detection in Persian tweets | 8,000 Persian tweets with emotional labels | Machine learning and deep learning techniques with various feature sets | 80.82% accuracy in sarcasm detection |
| **Golazizian** | Irony detection in Persian tweets | 4,339 manually labeled Persian tweets | BiLSTM network with attention mechanism | 83.1% accuracy in irony detection |
| **Hajiabdollah** | Improving sentiment analysis by incorporating sarcasm detection | 8,000 Persian tweets with emotional labels | Classifiers with sarcasm detection feature integration | Increased accuracy in sentiment analysis by incorporating irony detection |
| **Najafi** | Humor detection in Persian tweets | 14,946 automatically labeled Persian tweets | Fine-tuned transformer-based language model | 84.7% accuracy in humor detection |
| **Sharma** | Sentiment classification of short sentences using CNN with Word2Vec embeddings | Not specified, general short sentences | CNN with fine-tuned Word2Vec embeddings | Improved classification performance for short sentences |
| **Muhammad** | Sentiment analysis of hotel reviews in Indonesian using Word2Vec and LSTM | Indonesian hotel reviews | Word2Vec embeddings combined with LSTM networks | Significant improvement in sentiment classification accuracy |
| **Ouchene** | Sentiment analysis of Algerian tweets using FastText and LSTM | Algerian Arabic tweets | FastText embeddings with LSTM networks | Effective sentiment analysis in low-resource language |
| **Patel** | Rumor detection using hybrid deep learning approach | Not specified, rumor detection dataset | ELMo embeddings combined with CNN | Enhanced accuracy in rumor detection |
| **Farahani** | Persian language processing advancements with ParsBERT | Various Persian language tasks | Transformer-based ParsBERT model | Significant advancements in Persian NLP tasks |
| **Pires** | Multilingual capabilities of BERT | Various multilingual datasets | Analysis of multilingual BERT performance | Effective handling of multiple languages, including limited-resource languages |
| **Alkhalifa** | Humor detection in Arabic using pre-trained language models | Dataset of humorous tweets in Arabic | Fine-tuned pre-trained Arabic language models | CAMeLBERT-DA model achieved an F1-score of 72.11% |
| **Eke** | Sarcasm detection using context-based features | Twitter and Internet Argument Corpus datasets | Bi-LSTM with GloVe, BERT-based model, and feature fusion | High precision rates of 98.5% and 98.0% |
| **Shatnawi** | Humor detection in news headlines using BERT and Flair | SemEval-2020 public humor datasets | BERT-Flair-based ensemble model | RMSE of 0.51966 and accuracy of 0.62291 |
| **Annamoradnejad** | Humor detection and rating in short texts based on linguistic theory | 200,000 labeled short texts | Sentence embeddings with BERT and neural network | F1 scores of 0.982 and 0.869 in humor detection |
| **Sadjadi** | Measuring semantic similarity in Persian informal texts | FarSSiM dataset with 104 million Persian informal texts | Transformer-based FarSSiBERT model with novel tokenizer | Outperformed ParsBERT, laBSE, and multilingual BERT in similarity measurement |
| **Falakaflaki** | Formality style transfer in Persian, converting informal to formal text | Not specified | Fa-BERT2BERT model with consistency learning and dynamic weighting | Superior performance in BLEU, BERT score, Rouge-l, and other metrics |
| **Dashti** | Persian spelling correction integrating deep learning and phonetic analysis | Comprehensive spelling correction dataset | Fine-tuned language model with phonetic analysis | F1-Scores: 0.890 (real-word errors), 0.905 (corrections), 0.891 (non-word errors) |
| **Kebriaei** | Identification of hate and offensive language in Persian tweets | 38,000 Persian tweets with hate and offensive language annotations | Keyword-based data selection and crowdsourced lexicons | Effective bias mitigation with minimal impact on F1 score |
| **Vakili** | Advanced sentiment analysis for Persian Twitter content integrating Naive Bayes, rule-based model, and BERT | Persian Twitter content | Hybrid model combining Naive Bayes, rule-based, and BERT | Achieved 89% accuracy, outperforming BERT's 86% |

## Objectives

This research will employ quantitative research methodology to derive results. The approach involves the observation and analysis of authentic data, aimed at extracting and scrutinizing pertinent characteristics and variables, as well as evaluating the impact of each. Subsequently, the findings, along with the identification of optimal variables, will be reported based on the empirical data. The procedural steps for conducting this research include:

*A. Preliminary Study in the Required Fields*

A thorough and adequate examination of the definitions, terms, and existing literature relevant to the subject within the field is imperative.

*B. Conversion of Slang Texts into Formal*

The investigation focuses on delineating the definition of informal language and discerning its distinctions from formal language, in addition to elucidating the rules governing informal language. The goal is to devise a tool capable of transforming slang texts into formal expressions within the Persian language context. Furthermore, a comprehensive review of existing efforts in the domain of converting slang texts into formal language will be undertaken to inform and enrich the study. Given the Persian colloquial and formal grammar, a significant portion of colloquial words have been converted to formal ones using rule-based methods. Each of these methods not only affects the target words but also introduces errors in other words. Therefore, the effectiveness of these methods had to be evaluated through trial and error to ensure that if the generated error rate was negligible, the method would be chosen for converting colloquial words to formal ones.

*C. Sentiment Analysis of Persian Short Texts*

A comprehensive investigation within the domain of sentiment analysis is warranted, encompassing an exploration of definitions, levels, and practical applications within the field. Additionally, attention will be devoted to examining the characteristics of short texts and the associated challenges posed to machine learning algorithms. Furthermore, an inquiry into existing endeavors concerning sentiment analysis of both Persian and English short texts will be conducted to inform the research comprehensively.

*D. Data Gathering*

This research involves the collection of data from two distinct categories.

1. **Unlabeled Text:** To facilitate the conversion of conversational texts into formal expressions and to facilitate deep learning of short texts, the research will rely on two primary sources of data. The first source encompasses raw and untagged texts extracted from social networking platforms such as Instagram, Twitter, and Telegram, as well as subtitles from movies, which will serve as conversational data. The second source comprises texts sourced from Persian-language news agency websites, which will serve as formal texts for the study.
2. **Labeled Text:** For the implementation of the sentiment analysis model, supervised learning will be employed using tagged data derived from Instagram users' comments.

*E. Data Standardization*

To prepare the data for subsequent steps, a crucial pre-processing and standardization phase is imperative. Texts from social networks exhibit considerable complexity, attributable to diverse characters, hashtags, links, emoticons, and generally non-standard writing practices. Prior to utilization and processing, it is essential to standardize these data. While numerous basic natural language processing tools are available for the Persian language, their performance on conversational texts from social networks is often suboptimal. Consequently, there arises a necessity to implement specific equalization techniques tailored for these texts.

*F. Method*

Through a meticulous examination of the data and methodologies applied in addressing both the challenges of converting slang texts into formal expressions and analyzing the sentiments of short texts, the research will make informed decisions regarding the selection of methods. These methods may draw inspiration from prior research efforts or be entirely novel, tailored to the specific requirements of the current study. The decision-making process will be guided by a comprehensive understanding of the intricacies presented by the data and the specific objectives of the research.

*G. Implementation*

Based on the outcomes derived from the preceding stage, the chosen methods will be implemented utilizing the research data. Unlabeled data will be leveraged to train unsupervised models, while labeled data will be instrumental in training supervised models. Considering the substantial volume of data involved in this research, the utilization of appropriate hardware and up-to-date software packages is deemed necessary.

*H. Evaluation*

A pivotal stage in any system involves the evaluation and validation of the stated claims. In the context of this research, a comprehensive examination will be conducted to assess both the efficacy of converting slang texts into formal expressions and the accuracy of the sentiment analysis applied to short texts. This meticulous investigation aims to verify the reliability and effectiveness of the proposed methods within the defined scope of the research.

1. **Formal evaluation of the Proposed Tool:** Considering the scarcity of labeled data and the encompassing diversity within these datasets, the evaluation of this tool will pivot on the total number of converted words. This approach accounts for the comprehensive nature of the data, providing a broader assessment of the tool's performance in handling various linguistic nuances and expressions encountered in conversational texts.
2. **Short text Sentiment Classification Evaluation:** The model, constructed using the training data, is subsequently applied to the test data. The predictions generated by the model are then compared with the actual labels, and various evaluation criteria, including accuracy, precision, recall, and F-score, are computed. This meticulous assessment serves to gauge the performance of the model and validate its effectiveness in accurately predicting outcomes across the test dataset.
3. **PSC Evaluation for Short Text Sentiment Classification:** Furthermore, each classification involving slang texts and their corresponding converted expressions undergoes evaluation through the formalizer tool. This analysis aims to scrutinize the impact of the formalizer tool on the transformation process and assess its effectiveness in achieving the desired conversion of slang texts into formalized language.

## Data Section

The totality of the data used in this research is described in this section. The utilized data comprises two parts with labels and without labels. The labeled data are used to train the sentiment classification model.

### *A. Data Gathering*

Based on the conducted investigations, it appears that authentic conversational texts in the Persian language are scarcely available. However, there are corpora that includes complete sets of formal texts, such as the Hamshahri corpus containing several years' worth of news from this agency. While the news agency's content may seem comprehensive, a notable challenge lies in the fact that not all sentences within this corpus are strictly formal. Instances may arise where the Hamshahri news agency quotes colloquial expressions from individuals, introducing a layer of complexity to the categorization of formal and informal language within the dataset.

Given the inadequacy of Persian texts meeting the requirements of this treatise, it became imperative to procure colloquial and formal textual data from diverse sources. Based on the conducted surveys, the most suitable source for data collection emerged as the formal texts from news agencies. Numerous Farsi-language news agency websites are accessible, facilitating data extraction through web crawling. Additionally, recognizing that colloquial language is prevalent in everyday communication, social networks stand out as a rich source for collecting colloquial data. Many social networks offer programming interfaces that can be leveraged for this purpose.

1. **Data Crawling***:* Data crawling from the web involves extracting information and the structural elements of a web page, subsequently storing the acquired data in a database for individual retrieval of texts, images, videos, and other components. The initial phase of this information extraction process entails creating a robot (crawler) capable of recognizing the structure of a web page, executing parsing operations, and then storing the parsed data in a database. In this research, a custom-designed crawler was employed to fulfill these tasks.
2. **Social Networks APIs**: The process of data collection is significantly streamlined using interfaces, obviating the need to develop and program a crawler from scratch. Social networks offer interfaces to enhance profitability and foster business activities within their networks. Notably, tools such as the Telegram bot for Telegram and the Instagram API for Instagram have been employed for data collection in this research. The Telegram bot facilitates the collection and storage of diverse data from groups and channels, while the Instagram interface enables the gathering and storage of posts and comments from public Instagram networks.

### *B. Dataset*

The data collected for this research is presented in the Table 2. The dataset in the study is categorized into two main types: raw data and labeled data. This chapter will focus on the discussion of unlabeled data, while Table 3 will provide insights into the labeled data.

Table 2: Dataset statistic

| Data Subject | Counts |
|---|---|
| **Persian news agencies** | 10,000,000 |
| **Instagram Comments** | 6,000,000 |
| **Movies Subtitles** | 4,000,000 |
| **Telegram Groups** | 4,000,000 |
| **X (Twitter) Tweets** | 4,000,000 |
| **Instagram Comments** | 80,000 |
| **Sum** | 28,080,000 |

For unsupervised learning methods like Word2Vec, FastText, and BERT, which involve pre-training models, the efficacy and generalization of these models are highly dependent on the use of large datasets. The expansive datasets serve to impart a broad understanding of language patterns and nuances, enabling the models to capture a rich representation of the underlying linguistic structures. The use of substantial and diverse datasets is crucial in enhancing the robustness and performance of pre-trained models in various natural language processing tasks.

### *C. Data Preprocessing*

Before conducting any operations on textual data, preprocessing is essential to make the data usable. Text processing tools offer various functionalities such as equalization, stemming, tokenization, and sentence segmentation. Among these tools, Hazm is widely used for Persian text processing, particularly for formal texts. However, a separate tool was implemented for unifying conversational texts. The prevalence of non-standard words in everyday communication, especially in social networks, poses a challenge for machines to comprehend text content. These non-standard words can impact the performance of natural language processing tools, including machine translators, text summarizers, and text component taggers. To achieve the highest accuracy, text unification is crucial before any processing. Some of the equalization operations include Removal of links, IDs, and phrases related to social networks, such as retweets on Twitter.

- Separation of emoticons, punctuation, numbers, hashtags, English letters, and other characters,
- Unification of Persian and Arabic letters, as well as Persian and English numbers,
- Removal and correction of spaces and semi-spaces,
- Removal of Arabization, and
- Unification of polysyllabic words, such as "America" and its Persian equivalent.

These equalization operations collectively contribute to preparing the textual data for subsequent natural language processing tasks.

### *D. Constructing Term-Frequency (TF) Corpus*

Some text platforms predominantly contain colloquial text, while others are primarily formal; these can be utilized to establish word frequencies. For instance, raw comments from social networks can serve as colloquial texts, while raw texts from news agencies can be considered formal. However, a crucial consideration here is that certain texts, like comments on the Instagram social network, may lack entirely colloquial sentences, featuring formal expressions or even sentences in other languages or dialects. It is evident that such a corpus is more diverse and liberal, making these sentences and expressions more authentic and valid. After standardizing the corpora, the news corpus was designated as the formal corpus, and the aggregate of tweet corpora, Telegram messages, Instagram comments, and video subtitles were merged as the colloquial corpus. Subsequently, two corpora were created to quantify the frequency of formal and colloquial words. A corpus was specifically crafted under the title of "Pure Conversational." Pure Conversational is a corpus of words that adheres to specific criteria.

1. They must be present in the Conversational Corpus.
2. If they exist in the formal corpus, the number of repetitions in the colloquial corpus must be more than five times the number of repetitions in the formal corpus.

This coefficient was obtained by trial and error. The number of repetitions of words in each figure is available in the Table 3.

Table 3: Term Frequency (TF) corpus

| Frequency Corpus | Unique Words | Sum of Frequency |
|---|---|---|
| **Formal** | 570,092 | 726,905,260 |
| **Slang** | 1,465,468 | 70,236,528 |
| **Pure Slang** | 1,279,607 | 16,136,606 |

Also, the first 10 words of each figure and the number of repetitions in that figure are listed in the Table 4.

Table 4: The first 10 words of each figure and the number of repetitions in that figure

| Pure Slang | | Slang | | Formal | |
|---|---|---|---|---|---|
| 476,752 | یه | 1,626,397 | و | 33,567,634 | و |
| 214,569 | باشه | 1,230,548 | به | 27,528,074 | در |
| 204,356 | دیگه | 1,211,883 | که | 22,548,591 | به |
| 184,633 | داره | 964,338 | از | 17,483,954 | از |
| 177,246 | میشه | 878,994 | رو | 14,038,740 | این |
| 156,974 | اگه | 752,524 | این | 13,996,189 | که |
| 81,062 | میکنه | 714,285 | تو | 11,053,024 | با |
| 80,765 | بشه | 712,672 | من | 10,914,328 | است |

## PSC: Persian Slang Text Conversion to Formal

Many slang words can be formalized by examining the colloquial grammar rules explained in the previous chapter and studying the pure colloquial corpus through the application of a limited set of rules. These rules have been derived using the intelligent search method and visual observation to identify patterns and structures within colloquial language.

1. **Words Direct Conversion**: The number of repetitions of words in the Pure Conversational Corpus follows the Fig. 1. This chart follows a large head long tail distribution. By converting a few key words, many words in any colloquial sentence can be easily transformed. The 1000 most frequent words from the pure conversational corpus were directly and manually converted into their formal forms. By obtaining a pure colloquial corpus through the comparison of colloquial and formal data, and subsequently reviewing this corpus, interesting rules naturally come to mind.

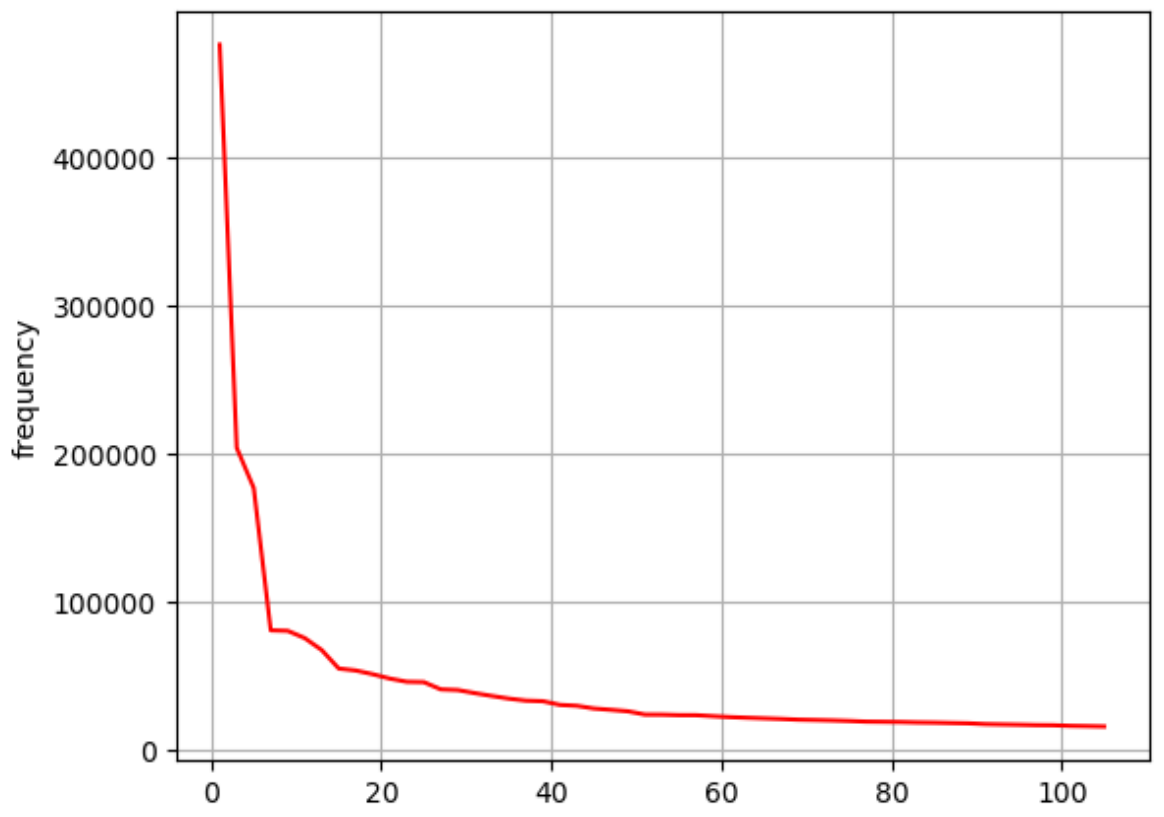

Fig. 1: The number of repetitions of words in the body of pure conversation.

- For example, after observing the words "*اون*" (oon), "*همون*" (hamoon), and "*خونه*" (khooneh), we conclude that by converting "*ون*" (oon) to "*ان*" (aan), several colloquial words can be transformed into their formal equivalents. However, it is inevitable to consider exceptions for each rule, such as the word "*خون*" (khoon) for this specific rule.
- **"و" to "را" Conversion** Strings that end with the letter "و" and are in the object sentence. First, it is checked that this word does not have more than 10 repetitions in the formal corpus. This is so that it does not become a mistake if the formal word existed with this form. After removing the "و" at the end of the word, the number of strings is checked in the formal body, and if they were more than 20, they become that substring plus "را". for example:
    - گلابیو : گلابی را
    - میزو : میز را
- **"ون" to "ان" Conversion** Strings ending in "ون". First, it is checked that this word does not have more than 10 repetitions in the official corpus. This is so that it does not become a mistake if the formal word existed with this form. Strings are checked for their number after removing "ون" at the end of the word in the formal body, and if they were more than 20, they become that substring plus "ان". For example:
    - خیابون : خیابان
    - صابون : صابون
- **Plural Words Conversion**: Strings ending in "ا" or "ای". First, it is checked that this word does not have more than 10 repetitions in the formal text. This is so that it does not become a mistake if the formal word existed with this form. After removing the "ا" at the end of the word, the number of strings is checked in the formal text, and if they were more than 20, they become that substring plus a half space plus "ها". For example:
    - خودکارا : خودکار ها
    - دارا : دارا
- **Repetition of Letters Conversion**: Sometimes users repeat one or more letters more when typing a word to emphasize or express a feeling. First, it is checked that this word does not have more than 10 repetitions in the formal text. This is so that it does not become a mistake if the formal word existed with this form. After correcting the repeated letters in the formal corpus, the number of strings is checked and if the number of repetitions is more than 20, they are converted. For example:
    - خخخخخخخخخخ : خ
    - پاییز : پاییز
    - مممممحححسسسسننننن : محسن
- **Colloquial Verbs Conversion**: In the last chapter, it was observed that all colloquial verbs, like formal verbs, have rules that can be converted by knowing this rule: prefix + past participle or participle + suffix. that by converting each of these parts into their formal forms, the entire verb can be converted. As:
    - بخون : بخوان
    - ترسوندم : ترساندم
    - رفتن : رفتن
    - ندونه : نداند

  Before the conversion, it is checked that if the number of repetitions of this word in the formal text is more than 10, this conversion will not be done. Also, a list was created for this section under the title of exception list. This list includes all colloquial and formal infinitives that cannot be converted directly. This is because all infinitives such as "*رفتن*" can be used in the third person plural form as well as in the form of a noun.
- **Possessive Pronouns Conversion**: All formal possessive pronouns have a rule that can be converted by knowing this rule. This is so that it does not become a mistake if the formal word existed with this form. Then it is checked if it ends with one of the possessive connected pronouns, they are divided into two substrings, the number of the first substring in the formal corpus is checked, and if they were more than 20, it becomes the first substring and the converted form of the second substring.

- خودکاراتون : خودکارهایتان
- هوام : هوایم
- قیافش : قیافه‌اش
- همسایمون : همسایه‌مان

- **The results of conversion of conversation to formal on pure slang corpus:** After implementing all the rules, each of these rules was executed on the pure slang corpus. In the table below, the ten most frequently converted words from each rule can be seen in the corpus of pure conversation. You can see the result. The number of unique converted words along with the total number of repetitions of these words for each rule is given in the table below.

Table 5 "ون" to "ان" conversion

| "ون" to "ان" Conversion | | |
|---|---|---|
| **Word** | **Converted** | **No. of Occurrence** |
| **آقایون** | آقایان | 5144 |
| **اقایون** | اقایان | 2379 |
| **یکیشون** | یکیشان | 2049 |
| **شیطون** | شیطان | 1898 |
| **برگردون** | برگردان | 1388 |
| **حالشون** | حالشان | 1360 |
| **اولشون** | اولشان | 1334 |
| **دمشون** | دمشان | 1299 |
| **زندگیشون** | زندگیشان | 1232 |
| **جفتشون** | جفتشان | 985 |

Table 6: "و" to "را" conversion

| "و" to "را" Conversion | | |
|---|---|---|
| **Word** | **Converted** | **No. of Occurrence** |
| **خودمو** | خودم را | 5017 |
| **داستانو** | داستان را | 3132 |
| **همینو** | همین را | 3087 |
| **دهنتو** | دهنت را | 2713 |
| **چیزو** | چیز را | 2606 |
| **اسمشو** | اسمش را | 2520 |
| **مملکتو** | مملکت را | 2291 |
| **اینجارو** | اینجا را | 2020 |
| **صدامو** | صدام را | 1944 |
| **حالمو** | حالم را | 1892 |

Table 7: Words direct conversion

| Words Direct Conversion | | |
|---|---|---|
| **Word** | **Converted** | **No. of Occurrence** |
| **یه** | یک | 476752 |
| **باشه** | باشد | 214569 |
| **دیگه** | دیگر | 204356 |
| **داره** | دارد | 184633 |
| **میشه** | می‌شود | 177246 |
| **اگه** | اگر | 156974 |
| **میکنه** | می‌کند | 81062 |
| **بشه** | بشود | 80765 |
| **واسه** | برای | 79297 |
| **آره** | بله | 75793 |

Table 8: Possessive pronouns conversion

| Possessive Pronouns Conversion | | |
|---|---|---|
| **Word** | **Converted** | **No. of Occurrence** |
| **پیداش** | پیدایش | 5694 |
| **شبتون** | شبتان | 5586 |
| **لطفتون** | لطفتان | 5212 |
| **اینجام** | اینجایم | 3438 |
| **بابام** | بابایم | 3432 |
| **دمتون** | دمتان | 3380 |
| **بابات** | بابایت | 3219 |
| **چشمات** | چشم‌هایت | 2926 |
| **شیش** | شیه‌اش | 2892 |
| **صبحتون** | صبحتان | 2755 |

Table 9: Repetition of letters conversion

| Repetition Of Letters Conversion | | |
|---|---|---|
| **Word** | **Converted** | **No. of Occurrence** |
| **خخخخ** | خ | 16972 |
| **خخخ** | خ | 14413 |
| **خخخخخ** | خ | 11548 |
| **خخخخخخ** | خ | 6561 |
| **خخخخخخخ** | خ | 3586 |
| **جووون** | جون | 2915 |
| **عااالی** | عالی | 2833 |
| **وااای** | وای | 2636 |
| **عاالی** | عالی | 2633 |
| **ههههه** | ه | 2628 |

Table 10: Colloquial verbs conversion

| Colloquial Verbs Conversion | | |
|---|---|---|
| **Word** | **Converted** | **No. of Occurrence** |
| **میکنی** | می‌کنی | 60881 |
| **نکنه** | نکند | 16695 |
| **می‌کنه** | می‌کند | 14846 |
| **میکردم** | می‌کردم | 14347 |
| **میخوان** | می‌خواهند | 11014 |
| **می‌خوام** | می‌خواهم | 10983 |
| **میزنم** | می‌زنم | 10759 |
| **میخواستم** | می‌خواستم | 10007 |
| **بکنه** | بکند | 8124 |
| **نمیده** | نمی‌دهد | 8065 |

Table 11: Plural words conversion

| Plural Words Conversion | | |
|---|---|---|
| **Word** | **Converted** | **No. of Occurrence** |
| **دخترا** | دخترها | 5607 |
| **بعضیا** | بعضی‌ها | 5373 |
| **حرفای** | حرف‌های | 5273 |
| **دوستای** | دوست‌های | 3371 |
| **بهترینها** | بهترین‌ها | 2728 |
| **چشمای** | چشم‌های | 2466 |
| **مردا** | مردها | 2208 |
| **دخترای** | دخترهای | 2204 |
| **خانوما** | خانوم‌ها | 1900 |
| **پستای** | پست‌های | 1771 |

Table 12: Comparison of rules statistics

| Rule | No. of UCW | % of UCW | No. of CW | % of CW |
|---|---|---|---|---|
| **Words Direct Conversion** | 1000 | 0.078 | 6514099 | 40.36 |
| **"و" to "ا" Conversion** | 22556 | 1.76 | 329726 | 0.043 |
| **"ون" to "ان" Conversion** | 3012 | 0.23 | 119466 | 0.74 |
| **Plural Words Conversion** | 31874 | 2.49 | 305020 | 1.89 |
| **Repetition Of Letters Conversion** | 65298 | 5.10 | 377351 | 2.33 |
| **Colloquial Verbs Conversion** | 4522 | 0.35 | 1021381 | 6.32 |
| **Possessive Pronouns Conversion** | 44288 | 3.46 | 574996 | 3.56 |
| **All Rules** | 172550 | 13.48 | 9242039 | 57.27 |

At the data section of our study, various types of rule-based conversion is shown in Table 5, Table 6, Table 7, Table 8, Table 9, Table 10, and Table 11. Also, statistics of rules comparison is shown in Table 12.

2. **Sentiment Labeled Data**: The dataset used in this research comprises 60,000 comments from the Instagram social network. Each comment has been labeled as positive, negative, or neutral by three users. The final labeling decision assigns the tag of the same class to comments that have at least two identical tags, achieving a consensus-based approach for classification. In the Table 13.

Table 13: Labeled data classification

| Label | Count | Sample |
|---|---|---|
| **Positive** | 25779 | درود بر بزرگ مرد اخلاق سینمای ایران وجهان |
| **Neutral** | 15366 | کمیاب ترین کتاب‌ها در کانال ما |
| **Negative** | 19657 | تأسف برای آن دسته از ادمهایی که انقدر بی ادب هستند |
| **Sum** | 60802 | درود بر بزرگ مرد اخلاق سینمای ایران وجهان |

To address data imbalance and ensure an equal number of instances for all classes, 15,000 sentences were randomly subsampled from both the positive and negative classes, resulting in an equal number for all three classes. Consequently, a total of 45,000 labeled data points were generated. In many research scenarios, a certain proportion of labeled data is randomly set aside for training and testing purposes. While some studies employ a three-way split (training, validation, and testing), others utilize the K-Fold method, although this is less common for larger datasets due to its computational complexity. In this research, as it is shown in Fig. 2 after performing pre-processing on all labeled and matched data, 70% of the data was allocated for training, with 15% each for validation and testing.

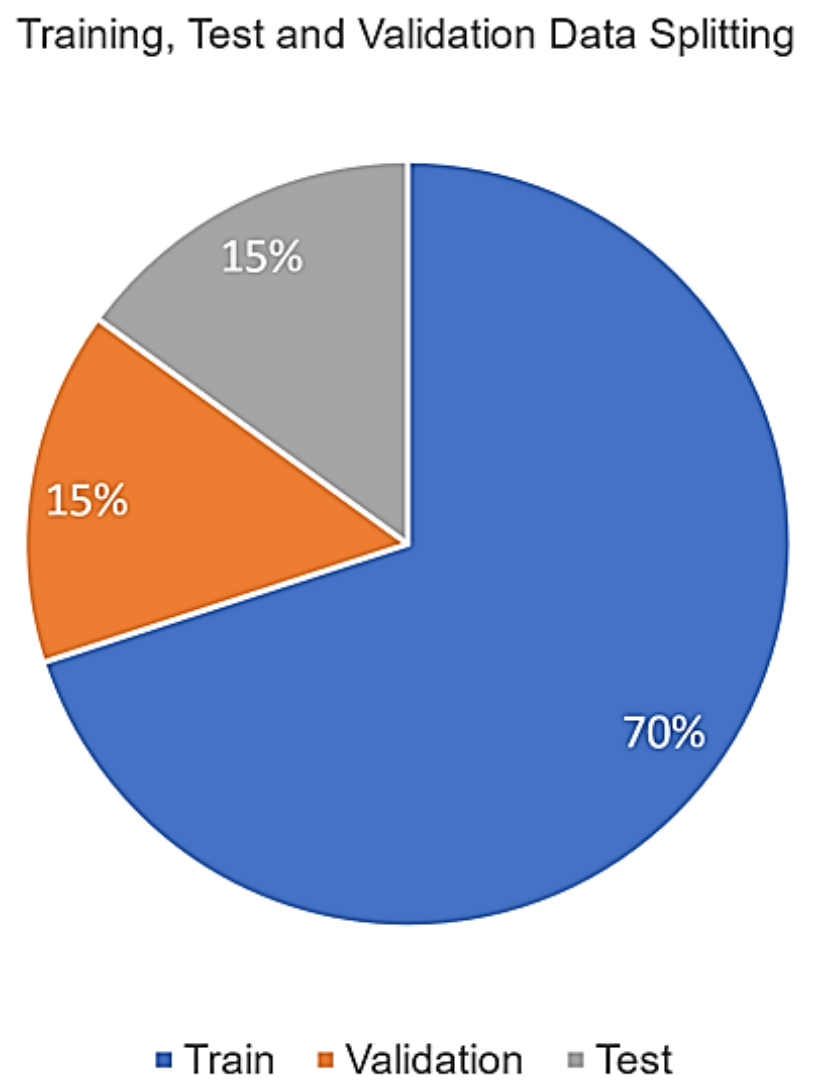

Fig. 2: Splitting rate of train, test and validation data.

## Results and Discussion

The algorithms were applied using labeled data, and their respective results are presented in Table 14. Additionally, the PSC metric has been employed to assess its performance on Sentiment classification.

Table 14: Algorithms results comparison

| Method | Precision | Accuracy | Recall | F1 |
|---|---|---|---|---|
| **Word2Vec+CNN** | 78.20 | 78.30 | 78.14 | 78.20 |
| **Word2Vec+CNN+PSC** | 78.90 | 78.91 | 78.83 | 78.86 |
| **word2Vec+LSTM** | 80.71 | 80.98 | 80.66 | 80.76 |
| **Word2Vec+LSTM+PSC** | 81.49 | 81.68 | 81.44 | 81.50 |
| **FastText+LSTM** | 81.21 | 81.12 | 81.10 | 81.11 |
| **FastText+LSTM+PSC** | **81.91** | **81.89** | **81.84** | **81.85** |
| **ELMo+CNN** | 77.61 | 77.87 | 77.41 | 77.11 |
| **ELMo+CNN+PSC** | 79.10 | 78.94 | 78.99 | 78.96 |
| **Parsbert** | 80.15 | 80.38 | 80.10 | 80.18 |
| **Parsbert+PSC** | 80.61 | 81.02 | 80.55 | 80.65 |
| **Bert_Multilingual** | 70.26 | 70.22 | 70.17 | 70.19 |

The Table 14 provide a comprehensive overview of the overall performance of the model across various deep learning methods. It is important to note that each cloud method involves numerous parameters, and the reported values represent the best-performing configurations in terms of accuracy. Upon analyzing the tables, the following conclusions can be drawn for deep learning methods:By using the PSC, the performance of all methods has improved, although this value is very low, and with the improvement of the PSC, the performance of the classifier also increases.

- The incorporation of PSC has demonstrated improvements in the performance of all methods, albeit the observed enhancement being relatively modest. Nevertheless, an increase in PSC corresponds to an improvement in classifier performance.
- The highest accuracy achieved is 81.91% using formal FastText vectors in conjunction with an LSTM network.
- Generally, deep processing methods exhibit markedly superior performance compared to machine learning methods. The FastText method outperforms the Word2Vec method, potentially because FastText considers word characters in addition to word embeddings.
- Although BERT-based methods were anticipated to yield superior results, the reduction in accuracy can be attributed to a mismatch in the text domains. The BERT models used in this research were pretrained on formal texts (Wikipedia and books), whereas the labeled data consists of conversational texts from social networks.

The PSC method has been evaluated using two parameters:

a. By counting the number of words converted in the pure colloquial corpus.

b. The effectiveness of this method is one of the best previous sentiment analysis methods.

The implementation of deep learning methods involved numerous hyperparameters. Initially, approximate values for most of these hyperparameters were set based on prior research. Subsequently, through a process of trial and error, the optimal value for each hyperparameter was sought in various iterations to achieve maximum accuracy. The early stopping method was employed to halt the training of the model.

If, after several training sessions, the error and accuracy of the model predictions for the validation data showed no improvement, the training process was terminated. Fig. 3 and Fig. 4 are illustrate the prediction error and accuracy of the model concerning repetitions in the training and validation data using the **FastText+LSTM+PSC** hybrid method.

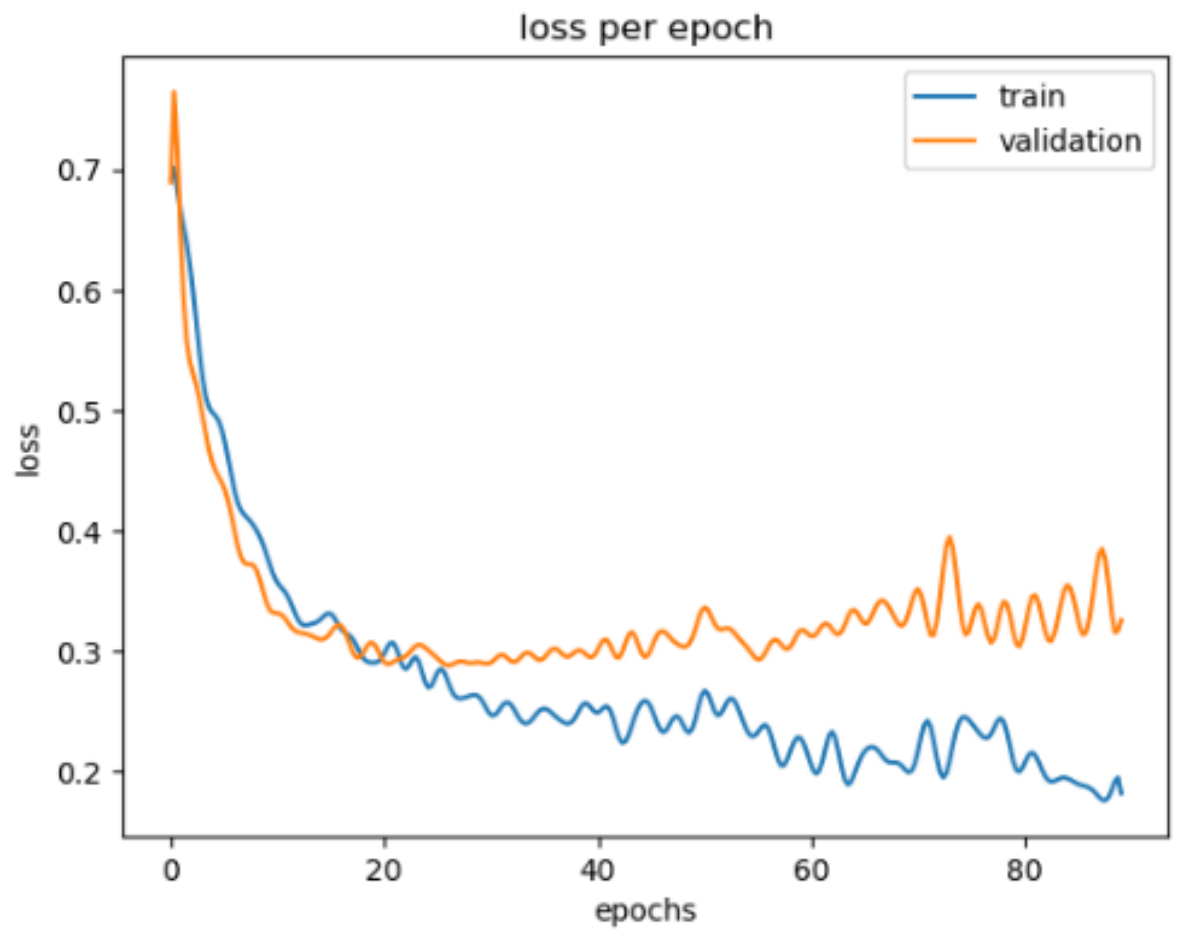

Fig. 3: Loss function.

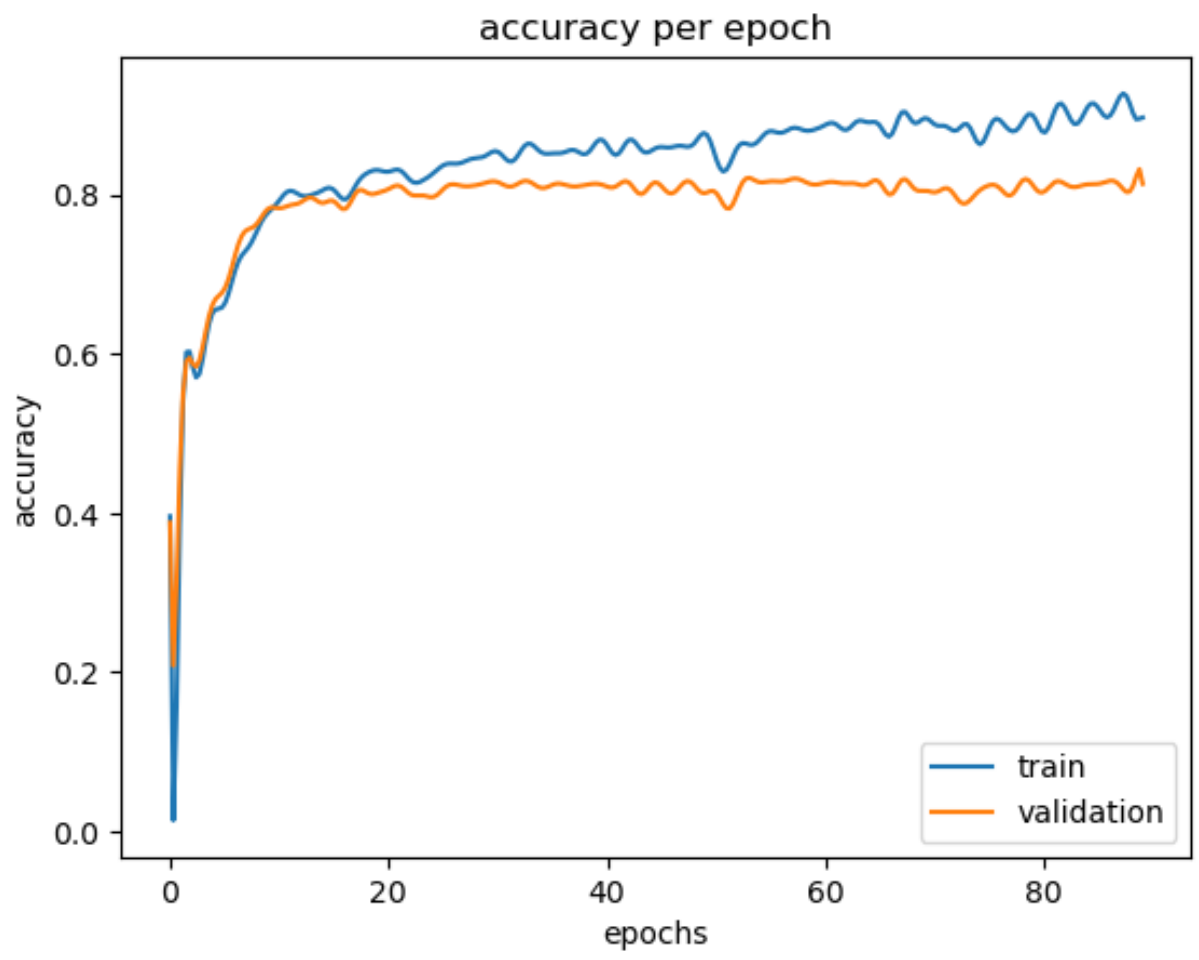

Fig. 4: Accuracy function.

After several iterations, the model's error increases for the validation data while decreasing for the training data, indicating a scenario of overfitting.

This phenomenon highlights the need for careful consideration and fine-tuning of hyperparameters to achieve optimal model performance.

## Conclusion

Today, with the increasing utilization of social networks by users, a substantial amount of valuable data is generated for analysis within these networks. Due to the diverse user base encompassing various tastes and age groups, the prevalence of colloquial language, along with abbreviations and numerous spelling and writing mistakes, has significantly grown. The evident lack of basic and advanced conversational language processing tools in the Persian language became a focal point in this research.

The aim was to enhance sentiment analysis by addressing two primary challenges present in textual data from social networks in Persian: shorthand writing and the use of colloquial expressions. The richness and comprehensiveness of the data used in the research play a pivotal role in ensuring the accuracy and thoroughness of the study. The dataset incorporated both labeled and unlabeled data. Labeled data comprised 60,000 sentences with three sentiment classes (positive, negative, and neutral) from the Instagram social network. Unlabeled data consisted of over 10 million sentences from social networks including Instagram, Twitter, and Telegram, representing conversational data, along with over 10 million texts from various news agencies, serving as formal data.

In the scope of converting slang texts into formal language, limited prior research has been conducted. The perceived lack of attention to the importance of this area and its inherent complexity may contribute to this gap. This research addressed this challenge by presenting a hybrid approach involving statistical and rule-based methods for converting colloquial texts into formal ones.

Three forms of data: formal, conversational, and pure conversation were created using the collected data, forming the foundation for the slang-to-formal conversion method. An analysis of the pure colloquial corpus revealed that implementing several rules can successfully convert a significant portion of slang words into formal ones. Following the applied checks, 57.2% of pure colloquial words were successfully converted using the implemented rules.

This research also focused on pre-training models using unlabeled data, occasionally leveraging existing pre-trained models like ParsBERT. Subsequently, a model was implemented to classify the sentiment of Persian short texts using labeled data, achieving a maximum accuracy of 81.9%.

## Future Works

As mentioned, there are several avenues for further exploration and enhancement in the field of converting slang texts into formal language, as well as in deep learning for short texts. The following points outline potential areas for continued research:

1. **Expand and Refine Rules:**

- Implement additional rules to enhance the formalization process, increasing the accuracy of current tools and rules.
- Consider contextual nuances, such as word position in a sentence or employing n-grams, to address errors in word conversion. For example, differentiating between "Selling their blood" and "Shed their blood."
- Explore sequence-to-sequence methods for sentence-level conversion, going beyond word-by-word transformation.

2. **Error Correction and Feature Preservation:**

- Investigate potential errors introduced by formalizing rules, especially in emotion-related texts. For instance, rules removing letter repetitions may inadvertently discard valuable features for emotion classification.
- Modify specific rules to improve the accuracy of emotion classification and ensure that important features are retained.

3. **Utilize NLP Tools for Improved Classification:**

- Leverage basic natural language processing tools, such as part-of-speech tagging and noun entity recognition, to enhance classifier performance.
- Assign higher weights to adjectives by incorporating part-of-speech tagging into feature vectors.
- Use noun entity recognition to remove nouns, allowing the model to grasp emotions from sentence style and context rather than learning specific nouns.

4. **Train Advanced Models in Persian:**

- Develop and train advanced models in Persian, akin to pre-trained models based on BERT available in English and other languages.
- Utilize large conversational datasets to train new models that could improve performance across various tasks, including emotion classification.
- Combine existing tagged data in Persian with additional datasets like Arman data to create a more robust model, especially in detecting allusions.

Continuing research in these areas could contribute to the refinement and advancement of tools and models, addressing challenges and optimizing performance in the analysis of emotions and the conversion of colloquial texts into formal language in the Persian language context.

## Author Contributions

Each author role in the research participation must be mentioned clearly.

Example:

M. Khazeni proposed the problem in the Persian NLP domain and designed the research roadmap. M. Khazeni Crawled the data from scratch. M. Khazeni, M. Heydari carried out the data analysis. M. Khazeni, M. Heydari interpreted the results and wrote the manuscript. A. Albadvi Supervised the entire study.

## Acknowledgment

The authors gratefully acknowledge the support and guidance of Professor Amir Albadvi for his work and supervision on proposing the research problem.

## Conflict of Interest

The authors declare no potential conflict of interest regarding the publication of this work. In addition, the ethical issues including plagiarism, informed consent, misconduct, data fabrication and, or falsification, double publication and, or submission, and redundancy have been completely witnessed by the authors.

## Abbreviations

In the final section of the article, abbreviations and their corresponding full forms are provided.

| | |
|---|---|
| *NLP* | Natural Language Processing |
| *LSTM* | Long-Short Term Memory |
| *BiLSTM* | Bidirectional Long-Short Term Memory |
| *CNN* | Convolutional Neural Networks |
| *BLEU* | Bilingual Evaluation Understudy |
| *BERT* | Bidirectional Encoder Representations from Transformers |
| *ELMO* | Embeddings from Language Model |
| *PSC* | Persian Slang Text Convertor |
| *MSA* | Modern Standard Arabic |
| *PLMs* | Pre-trained Language Models |
| *RMSE* | Root Mean Squared Error |
| *GloVe* | Global Vectors for Word Representation |
| *SVM* | Support Vector Machine |
| *MLP* | Multilayer Perceptron |
| *TF* | Term-Frequency |

## Biographies

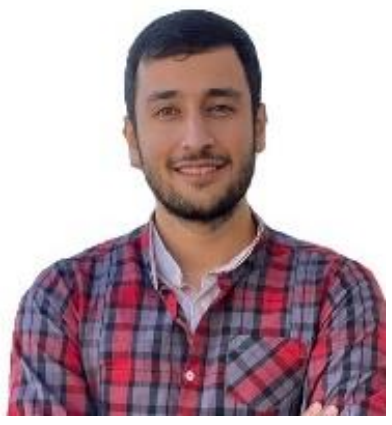

**Mohsen Khazeni** received his B.Sc degree in Computer Software Engineering, Iran University of Science and Technology and received his M.Sc. degree in Information Technology Engineering, Tarbiat Modares University, Tehran, Iran. His research interests are Natural Language Processing, Deep Learning, and Social Network Analysis.

- Email: m.khazeni@modares.ac.ir
- ORCID: NA
- Web of Science Researcher ID: NA
- Scopus Author ID: NA
- Homepage: https://researchgate.net/profile/Mohsen-Khazeni-2

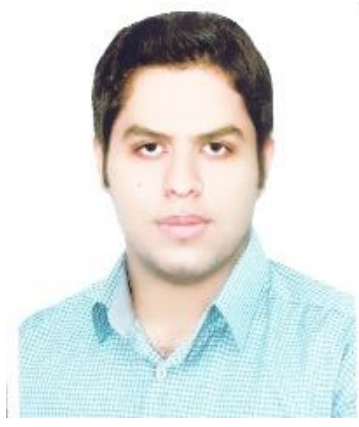

**Mohammad Heydari** received his B.Sc. degree in Computer Software Engineering, Technical and Vocational University, Tehan, Iran and received his M.Sc. degree in Information Technology Engineering, Tarbiat Modares University, Tehran, Iran. His research interests include Machine Learning, Big Data Engineering, and Graph Neural Networks.

- Email: m_heydari@modares.ac.ir
- ORCID: 0000-0002-7650-5924
- Web of Science Researcher ID: KZU-4848-2024
- Scopus Author ID: NA
- Homepage: https://scholar.google.com/citations?user=XBp6ipEAAAAJ&hl=en

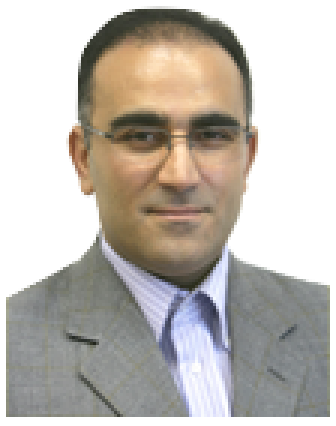

**Amir Albadvi** is Full Professor of Information Systems. He received his Ph.D. from London School of Economics (LSE) in London and eventually earned his spot as technology thought leader for extra-large technology transformation projects with over 20 years of experiences in IT transformation and e-Strategy. After 10 years career as a change professional and winning the prestigious IT Deployment Award for his contribution in IT implementation, he decided it was time for a change of scenery (and weather) and moved to Beautiful British Columbia, Vancouver where he was offered visiting professor position at UBC and Victoria University. In addition, he focused on start-up ecosystem in the region, established Parallax Solutions Enterprise for technology and management consulting. Dr. Albadvi is now involved in another initiative in design thinking and technology-based innovation as team DNA named "Albadvi & Associates". A novel platform to promote social entrepreneurship among young generations.

- Email: albadvi@modares.ac.ir
- ORCID: 0000-0002-7758-9920
- Web of Science Researcher ID: NA
- Scopus Author ID: NA
- Homepage: https://modares.ac.ir/~albadvi